\documentclass[letterpaper,11pt]{article}
\usepackage[margin=1in]{geometry}

\usepackage[utf8]{inputenc} 
\usepackage[T1]{fontenc}    
\usepackage{hyperref}       
\usepackage{url}            
\usepackage{booktabs}       
\usepackage{amsfonts}       
\usepackage{nicefrac}       
\usepackage{microtype}      

\usepackage{graphicx}
\usepackage{amssymb,amsmath,amsthm,amsfonts}
\usepackage{dsfont}
\usepackage{hyperref}
\usepackage{url}
\usepackage{stmaryrd}
\usepackage[noend]{algpseudocode}
\usepackage{algorithm}
\usepackage{float}

\DeclareMathAlphabet{\varmathbb}{U}{bbold}{m}{n}

\renewcommand{\d}{{\rm d}}
\renewcommand{\P}{{\mathbb{P}}}
\newcommand{\E}{{\mathbb{E}}}

\newcommand{\B}{{\rm B}}

\newcommand{\iif}{\Longleftrightarrow}

\newtheorem{theorem}{Theorem}

\def\ind{{\mathbf{1}}}
\newcommand{\trans}[1]{{#1}^\intercal}

\newcommand{\LineIf}[2]{ \State \algorithmicif\ {#1}\ \algorithmicthen\ {#2} }
\newcommand{\LineFor}[2]{ \State \algorithmicfor\ {#1}\ \algorithmicdo\ {#2} }
\newcommand{\LineElsIf}[2]{ \State \textbf{else if}\ {#1}\ \algorithmicthen\ {#2} }
\newcommand{\LineElse}[1]{ \State \textbf{else}\ {#1} }

\makeatletter
\renewcommand*\env@matrix[1][\arraystretch]{%
  \edef\arraystretch{#1}%
  \hskip -\arraycolsep
  \let\@ifnextchar\new@ifnextchar
  \array{*\c@MaxMatrixCols c}}
\makeatother

\algdef{SE}[SUBALG]{Indent}{EndIndent}{}{\algorithmicend\ }%
\algtext*{Indent}
\algtext*{EndIndent}

\makeatletter
\let\@fnsymbol\@arabic
\makeatother

\begin{document}

\title{Fast Randomized Semi-Supervised Clustering}



\author{Alaa Saade\thanks{
Laboratoire de Physique Statistique (CNRS UMR-8550),
  PSL Universit\'es \& \'Ecole Normale Sup\'erieure, 75005
  Paris (e-mail: \texttt{\{alaa.saade,florent.krzakala\}@ens.fr})
}\and Florent Krzakala$^{1,}$\footnote{
Sorbonne Universit\'es, UPMC Univ. Paris 06
}\and Marc Lelarge\thanks{
INRIA and \'Ecole Normale Sup\'erieure Paris, France (e-mail: \texttt{marc.lelarge@ens.fr})
}\and Lenka Zdeborov\'a\thanks{
  Institut de Physique Th\'eorique CEA Saclay and CNRS, France (e-mail: \texttt{lenka.zdeborova@gmail.com})
}
}
\date{}



\maketitle

\begin{abstract}
  We consider the problem of clustering partially labeled data from a minimal number of randomly chosen pairwise comparisons between the items. We introduce an efficient local algorithm based on a power iteration of the non-backtracking operator and study its performance on a simple model. For the case of two clusters, we give bounds on the classification error and show that a small error can be achieved from $O(n)$ randomly chosen measurements, where $n$ is the number of items in the dataset. Our algorithm is therefore efficient both in terms of time and space complexities. We also investigate numerically the performance of the algorithm on synthetic and real world data.
\end{abstract}

\section{Introduction}

Similarity-based clustering aims at classifying data points into homogeneous groups based on some measure of their resemblance. The problem can be stated formally as follows: given $n$ items $\{x_i\}_{i\in[n]}\in \mathcal{X}^n$, and a symmetric similarity function $s : \mathcal{X}^2 \to \mathbb{R}$, the aim is to cluster the dataset from the knowledge of the pairwise similarities $s_{ij} := s(x_{i},x_{j})$, for $1\leq i < j \leq n$. This information is usually represented in the form of a weighted similarity graph $G = (V,E)$ where the nodes represent the items of the dataset and the edges are weighted by the pairwise similarities. Popular choices for the similarity graph are the fully-connected graph, or the $k$-nearest neighbors graph, for a suitable $k$ (see e.g. \cite{Tutorial} for a review). Both choices, however, require the computation of a large number of pairwise similarities, typically $O(n^2)$. For large datasets, with $n$ in the millions or billions, or large-dimensional data, where computing each similarity $s_{ij}$ is costly, the complexity of this procedure is often prohibitive, both in terms of computational and memory requirements.

It is then natural to ask whether it is really required to compute as many as $O(n^2)$ similarities to accurately cluster the data. In the absence of additional information on the data, a reasonable alternative is to compare each item to a small number of other items in the dataset, chosen uniformly at random. Random subsampling methods are a well-known means of reducing the complexity of a problem, and they have been shown to yield substantial speed-ups in clustering \cite{drineas1999clustering} and low-rank approximation \cite{achlioptas2007fast,halko2011finding}. In particular, \cite{saade2016clustering} recently showed that an unsupervised spectral method based on the principal eigenvectors of the non-backtracking operator of \cite{krzakala2013spectral} can cluster the data better than chance from only $O(n)$ similarities. 

In this paper, we build upon previous work by considering two variations motivated by real-world applications. The first question we address is how to incorporate the knowledge of the labels of a small fraction of the items to aid clustering of the whole dataset, resulting in a more efficient algorithm.  
This question, referred to as semi-supervised clustering, is of broad practical interest \cite{basu2002semi,zhu2002learning}. For instance, in a social network, we may have pre-identified individuals of interest, and we might be looking for other individuals sharing similar characteristics. In biological networks, the function of some genes or proteins may have been determined by costly experiments, and we might seek other genes or proteins sharing the same function. More generally, efficient human-powered methods such as crowdsourcing can be used to accurately label part of the data \cite{karger2011iterative,karger2013efficient}, and we might want to use this knowledge to cluster the rest of the dataset at no additional cost. 

The second question we address is the number of randomly chosen pairwise
similarities that are needed to achieve a given classification
error.  Previous work has mainly focused on two related, but different
questions. One line of research has been interested in exact recovery,
i.e. how many measurements are necessary to exactly cluster the
data. Note that for exact recovery to be possible, it is necessary to
choose at least $O(n\log n)$ random measurements for the similarity
graph to be connected with high probability. On simple models,
\cite{abbe2014decoding,yun2015optimal,hajek2015achieving} showed that
this scaling is also sufficient for exact recovery. On the sparser end
of the spectrum,
\cite{LMX2013,HLM2012,saade2015spectral,saade2016clustering} have
focused on the detectability threshold, i.e. how many measurements are
needed to cluster the data better than chance. On simple models, this
threshold is typically achievable with $O(n)$ measurements only. While
this scaling is certainly attractive for large problems, it is
important for practical applications to understand how the expected
classification error decays with the number of measurements.

To answer these two questions, we introduce a highly efficient, local algorithm based on a power iteration of the non-backtracking operator. For the case of two clusters, we show on a simple but reasonable model that the classification error decays exponentially with the number of measured pairwise similarities, thus allowing the algorithm to cluster data to arbitrary accuracy while being efficient both in terms of time and space complexities. 
We demonstrate the good performance of this algorithm on both synthetic and real-world data, and compare it to the popular label propagation algorithm \cite{zhu2002learning}.  

\section{Algorithm and guarantee}
\label{sec:algo_guarante}


\subsection{Algorithm for $2$ clusters}
\label{subsec:algo_2_clusters}
Consider $n$ items $\{x_i\}_{i\in[n]}\in \mathcal{X}^n$ and a symmetric similarity function $s : \mathcal{X}^2 \to \mathbb{R}$. The choice of the similarity function is problem-dependent, and we will assume one has been chosen. For concreteness, $s$ can be thought of as a decreasing function of a distance if $\mathcal{X}$ is an Euclidean space. The following analysis, however, applies to a generic function $s$, and our bounds will depend explicitly on its statistical properties. We assume that the true labels $(\sigma_{i}=\pm1)_{i\in\mathcal{L}}$ of a subset $\mathcal{L}\subset[n]$ of items is known. Our aim is to find an estimate $(\sigma_{i})_{i\in[n]}$ of the labels of all the items, using a small number of similarities. 
More precisely, let $E$ be a random subset of all the ${n \choose 2}$ possible pairs of items, containing each given pair $(ij)\in[n]^{2}$ with probability $\alpha/n$, for some $\alpha > 0$. $E$
We compute only the $O(\alpha n)$ similarities $(s_{ij}:=s(x_{i},x_{j}))_{(ij)\in E}$ of the pairs thus chosen.

From these similarities, we define a weighted similarity graph $G=(V,E)$ where the vertices $V =[n]$ represent the items, and each edge $(ij)\in E$ carries a weight $w_{ij}:=w(s_{ij})$, where $w$ is a weighting function. Once more, we will consider a generic function $w$ in our analysis, and discuss the performance of our algorithm as a function of the choice of $w$. In particular, we show in section \ref{subsec:model_and_guarantee} that there is an optimal choice of $w$ when the data is generated from a model.
However, in practice, the main purpose of $w$ is to center the similarities, i.e. we will take in our numerical simulations $w(s) = s - \bar{s}$, where $\bar{s}$ is the empirical mean of the observed similarities. The necessity to center the similarities is discussed in the following. 
Note that the graph $G$ is a weighted version of an Erd\H{o}s-R\'enyi random graph with average degree $\alpha$, which controls the sampling rate: a larger $\alpha$ means more pairwise similarities are computed, at the expense of an increase in complexity. 
Algorithm \ref{alg:2_clusters} describes our clustering procedure for the case of $2$ clusters.
We denote by $\partial i$ the set of neighbors of node $i$ in the graph $G$, and by $\vec{E}$ the set of directed edges of $G$.

\begin{algorithm}
\caption{Non-backtracking local walk ($2$ clusters)}
\label{alg:2_clusters}
\begin{algorithmic}[1]
\Require{$n\,,\mathcal{L}\,,(\sigma_i = \pm 1)_{i\in\mathcal{L}}\,,E\,,(w_{ij})_{(ij)\in E}\,,k_{\rm max}$}
\Ensure{Cluster assignments $(\hat{\sigma}_i)_{i\in[n]}$ }
\State \textbf{Initialize} the messages $v^{(0)}=(v_{i\to j}^{(0)})_{(i\to j)\in \vec{E}}$\ 
\Indent
\ForAll{$(i\to j)\in \vec{E}$} \unskip 
\LineIf{$i\in\mathcal{L}$}{$v_{i\to j}^{(0)} \leftarrow \sigma_i$}
\LineElse{$v_{i\to j}^{(0)}\leftarrow \pm 1$ uniformly at random}
\EndFor
\EndIndent
\State \textbf{Iterate} for $k=1,\dots,k_{\rm max}$
\Indent
\LineFor{$(i\to j)\in \vec{E}$}{$v_{i\to j}^{(k)}\leftarrow \sum_{l\in\partial i\backslash j}w_{il} v_{l\to i}^{(k-1)}$}
\EndIndent
\State \textbf{Pool} the messages 
\Indent
\LineFor{$i\in [n]$}{$\hat{v}_{i} \leftarrow \sum_{l\in\partial i} w_{il}v_{l\to i}^{(k_{\rm max})}$}
\EndIndent
\State \textbf{Output} the assignments 
\Indent
\LineFor{$i\in [n]$}{$\hat{\sigma}_{i}\leftarrow {\rm sign}(\hat{v}_{i})$}
\EndIndent
\end{algorithmic}
\end{algorithm}
This algorithm can be thought of as a linearized version of a belief propagation algorithm, that iteratively updates messages on the directed edges of the similarity graph, by assigning to each message the weighted sum of its incoming messages. More precisely, algorithm \ref{alg:2_clusters} can be observed to approximate the leading eigenvector of the non-backtracking operator $\B$, whose elements are defined, for $(i\to j),(k\to l)\in\vec{E}$, by 
\begin{equation}
\label{def_B}
\B_{(i\to j),(k\to l)} := w_{kl}\,\ind(i=l)\ind(k\neq j)\, . 
\end{equation}
It is therefore close in spirit to the unsupervised spectral methods introduced by \cite{krzakala2013spectral,saade2015spectral,saade2016clustering}, which rely on the computation of the principal eigenvector of $\B$. On sparse graphs, methods based on the non-backtracking operator are known to perform better than traditional spectral algorithms based e.g. on the adjacency matrix, or random walk matrix of the sparse similarity graph, which suffer from the appearance of large eigenvalues with localized eigenvectors (see e.g. \cite{krzakala2013spectral,zhang2016robust}). In particular, we will see on the numerical experiments of section~\ref{sec:numerical_simulations} that algorithm~\ref{alg:2_clusters} outperforms the label propagation algorithm, based on an iteration of the random walk matrix.

However, in contrast with the past spectral approaches based on $\B$, our algorithm is local, in that the estimate $\hat{\sigma}_{i}$ for a given item $i\in[n]$ depends only on the messages on the edges that are at most $k_{\rm max}+1$ steps away from $i$ in the graph $G$. This fact will prove essential in our analysis. Indeed, we will show that in our semi-supervised setting, a finite number of iterations (independent of $n$) is enough to ensure a low classification error. On the other hand, in the unsupervised setting, we expect local algorithms not to be able to find large clusters in a graph, a limitation that has already been highlighted on the related problems of finding large independent sets on graphs \cite{gamarnik2014limits} and community detection \cite{kanade2014global}. On the practical side, the local nature of algorithm \ref{alg:2_clusters} leads to a drastic improvement in running time. Indeed, in order to compute the leading eigenvector of $\B$, a number of iterations $k$ scaling with the number $n$ of items is required \cite{blm15}. Here, on the contrary, the number of iterations stays independent of the size $n$ of the dataset.

\subsection{Model and guarantee}
\label{subsec:model_and_guarantee}
To evaluate the performance of algorithm \ref{alg:2_clusters}, we consider the
following 
semi-supervised variant of the labeled stochastic block model \cite{HLM2012}, a popular benchmark for graph clustering.
We assign $n$ items to $2$ predefined clusters of same average size $n/2$, by drawing for each item $i \in [n]$ a cluster label $\sigma_{i}\in\{\pm 1\}$ with uniform probability $1/2$. We choose uniformly at random $\eta n$ items to form a subset $\mathcal{L}\subset [n]$ of items whose label is revealed, so that $\eta$ is the fraction of labeled data.
Next, we choose which pairs of items will be compared by generating an Erd\H{o}s-R\'enyi random graph $G=(V=[n],E)\in {\cal G}(n,\alpha/n)$, for some constant $\alpha > 0$, independent of $n$.
We will assume that the similarity $s_{ij}$ between two items $i$ and $j$ is a random variable depending only on the labels of the items $i$ and $j$. More precisely, we consider the symmetric model
\begin{align}
\label{symmetric_model}
\P(s_{ij} = s|\sigma_{i}\,,\sigma_{j}) = 
\left\{
\begin{aligned}
&p_{\rm in}(s)&\text{ if }&\sigma_i=\sigma_j\, , \\
&p_{\rm out}(s)&\text{ if }&\sigma_i\neq\sigma_j\, ,
\end{aligned}
\right.
\end{align}
where $p_{\rm in}$ (resp. $p_{\rm out}$) is the distribution of the similarities between items within the same cluster (resp. different clusters). 
The properties of the weighting function $w$ will determine the
performance of our algorithm through the two following
quantities. Define $2\Delta(w) :=
\E\left[w_{ij}|\sigma_{j}=\sigma_{j}\right] -
\E\left[w_{ij}|\sigma_{j}\neq\sigma_{j}\right]$, the difference in
expectation between the weights inside a cluster and between different
clusters. Define also $\Sigma(w)^{2} := \E\left[w^{2}\right]$, the
second moment of the weights. 
Our first result (proved in section \ref{sec:analysis_density_evolution}) is concerned with what value of $\alpha$ is required to improve the initial labeling with algorithm \ref{alg:2_clusters}.
\begin{theorem}
\label{proposition}
Assume a similarity graph $G$ with $n$ items and a labeled set $\mathcal{L}$ of size $\eta n$ to be generated from the symmetric model (\ref{symmetric_model}) with $2$ clusters. 
Define $\tau(\alpha,w) := \frac{\alpha\Delta(w)^{2}}{\Sigma(w)^{2}}$. If $\Delta(w)>0$, then there exists a constant $C>0$ such that the estimates $\hat{\sigma}_i$ from $k$ iterations of algorithm \ref{alg:2_clusters} achieve 
\begin{equation}
\label{proposition_cantelli_bound}
\frac{1}{n}\sum_{i=1}^{n} \P(\sigma_{i}\neq\hat{\sigma}_i) \leq 1-r_{k+1} + C\frac{\alpha^{k+1}\log n}{\sqrt{n}}\, ,
\end{equation}
where $r_0 = \eta^{2}$ and for $0\leq l\leq k$,
\begin{equation}
\label{proposition_rec_r}
\displaystyle r_{l+1} = \frac{\tau(\alpha,w) r_{l}}{1 + \tau(\alpha,w) r_{l}}\, .
\end{equation}
\end{theorem}

To understand the content of this bound, we consider the limit of a large number of items $n\to\infty$, so that the last term of (\ref{proposition_cantelli_bound}) vanishes.
Note first that if $\tau(\alpha,w)>1$, then starting from any positive initial condition, $r_{k}$ converges to $(\tau(\alpha,w)-1)/\tau(\alpha,w)>0$ in the limit where the number of iterations $k\to\infty$. A random guess on the unlabeled points yields an asymptotic error of 
\begin{equation}
\label{random_guess}
\lim_{n\to \infty} \frac{1}{n}\sum_{i=1}^{n} \P(\sigma_{i}\neq\hat{\sigma}_i) = \frac{1-\eta}{2}\, ,
\end{equation}
so that a sufficient condition for algorithm \ref{alg:2_clusters} to improve the initial partial labeling, after a certain number of iterations $k(\tau(\alpha,w),\eta)$ independent of $n$, is
\begin{equation}
\label{bound_tau_improve_random_guess}
\tau(\alpha,w) > \frac{2}{1-\eta}\, .
\end{equation}
It is informative to compare this bound to known optimal asymptotic bounds in the unsupervised setting $\eta\to 0$.
Note first (consistently with \cite{LMX2013}) that there is an optimal choice of weighting function $w$ which maximizes $\tau(\alpha,w)$, namely 
\begin{equation}
\begin{aligned}
\label{optimal_w}
w^{*}(s) := \frac{p_{\rm in}(s) - p_{\rm out}(s)}{p_{\rm in}(s) + p_{\rm out}(s)}
\qquad\iif\qquad \tau(\alpha,w^{*}) = \frac{\alpha}{2}\int \d s\ \frac{(p_{\rm in}(s)-p_{\rm out}(s))^{2}}{p_{\rm in}(s)+p_{\rm out}(s)}\, ,
\end{aligned}
\end{equation}
which, however, requires knowing the parameters of the model. In the limit of vanishing supervision $\eta\to 0$, the bound (\ref{bound_tau_improve_random_guess}) guarantees improving the initial labeling if $\tau(\alpha,w^{*})>2+O(\eta)$. In the unsupervised setting ($\eta=0$), it has been shown by \cite{LMX2013} that if $\tau(\alpha,w^{*})<1$, no algorithm, either local or global, can cluster the graph better than random guessing. If $\tau(\alpha,w^{*})>1$, on the other hand, then a global spectral method based on the principal eigenvectors of the non-backtracking operator improves over random guessing \cite{saade2016clustering}. This suggests that, in the limit of vanishing supervision $\eta\to 0$, the bound (\ref{bound_tau_improve_random_guess}) is close to optimal, but off by a factor of $2$.
Note however that theorem \ref{proposition} applies to a generic weighting function $w$. In particular, while the optimal choice (\ref{optimal_w}) is not practical, theorem \ref{proposition} guarantees that algorithm \ref{alg:2_clusters} retains the ability to improve the initial labeling from a small number of measurements, as long as $\Delta(w)>0$. With the choice $w(s) = s-\bar{s}$ advocated for in section \ref{subsec:algo_2_clusters}, we have $2\Delta(w) = \E\left[s_{ij}|\sigma_{j}=\sigma_{j}\right] - \E\left[s_{ij}|\sigma_{j}\neq\sigma_{j}\right]$. Therefore algorithm \ref{alg:2_clusters} improves over random guessing for $\alpha$ large enough if the similarity between items in the same cluster is larger in expectation than the similarity between items in different clusters, which is a reasonable requirement. Note that the hypotheses of theorem \ref{proposition} do not require the weighting function $w$ to be centered. However, it is easy to check that if $\E[w]\neq 0$, defining a new weighting function by $w^{\prime}(s):=w(s)-\E[w]$, we have $\tau(\alpha,w^{\prime})>\tau(\alpha,w)$, so that the bound (\ref{proposition_cantelli_bound}) is improved. 

While theorem \ref{proposition} guarantees improving the initial clustering from a small sampling rate $\alpha$, it provides a rather loose bound on the expected error when $\alpha$ becomes larger. The next theorem addresses this regime. A proof is given in section \ref{proof_theorem}.

\begin{theorem}
\label{theorem}
Assume a similarity graph $G$ with $n$ items and a labeled set $\mathcal{L}$ of size $\eta n$ to be generated from the symmetric model (\ref{symmetric_model}) with $2$ clusters. Assume further that the weighting function $w$ is bounded: $\forall s\in\mathbb{R}, |w(s)|\leq 1$. Define $\tau(\alpha,w) := \frac{\alpha\Delta(w)^{2}}{\Sigma(w)^{2}}$. If~$\alpha\Delta(w)>1$ and $\alpha\Sigma(w)^{2}>1$, then there exists a constant $C>0$ such that the estimates $\hat{\sigma}_i$ from $k$ iterations of algorithm \ref{alg:2_clusters} achieve 
\begin{equation}
\begin{aligned}
\label{theorem_exponenetial_bound}
\frac{1}{n}\sum_{i=1}^{n} \P(\sigma_{i}\neq\hat{\sigma}_i) \leq \exp{\left[-\frac{q_{k+1}}{4}\min\left(1,\frac{\Sigma(w)^{2}}{\Delta(w)}\right)\right]}
+ C\frac{\alpha^{k+1}\log n}{\sqrt{n}}\, ,
\end{aligned}
\end{equation}
where $q_0 = 2\eta^{2}$ and for $0\leq l\leq k$,
\begin{equation}
\label{recurrence_q}
\displaystyle q_{l+1} = \frac{\tau(\alpha,w) q_{l}}{1 + 3/2\max(1,q_{l})}\, .
\end{equation}
\end{theorem}
Note that by linearity of algorithm \ref{alg:2_clusters}, the condition $\forall s, |w(s)|\leq 1$ can be relaxed to $w$ bounded. 
It is once more instructive to consider the limit of large number of items $n\to\infty$. Starting from any initial condition, if $\tau(\alpha,w)<5/2$, then $q_{k}\underset{k\to \infty}{\longrightarrow} 0$ so that the bound (\ref{theorem_exponenetial_bound}) is uninformative. On the other hand, if $\tau(\alpha,w)>5/2$, then starting from any positive initial condition, $q_{k}\underset{k\to \infty}{\longrightarrow} \frac{2}{3}(\tau(\alpha,w) -1) > 0$. 
This bound therefore shows that on a model with a given distribution of similarities (\ref{symmetric_model}) and a given weighting function $w$, an error smaller than $\epsilon$ can be achieved from $\alpha n = O(n\log{1/\epsilon})$ measurements, in the limit $\epsilon\to 0$, with a finite number of iterations $k(\tau(\alpha,w),\eta,\epsilon)$ independent of $n$. We note that this result is the analog, for a weighted graph, of the recent results of \cite{cai2016inference} who show that in the stochastic block model, a local algorithm similar to algorithm \ref{subsec:algo_2_clusters} achieves an error decaying exponentially as a function of a relevant signal to noise ratio.   

\subsection{More than $2$ clusters}
\label{subsec:more_than_2_clusters}

Algorithm \ref{alg:q_clusters} gives a natural extension of our algorithm to $q>2$ clusters. In this case, we expect the non-backtracking operator $\B$ defined in equation (\ref{def_B}) to have $q-1$ large eigenvalues, with eigenvectors correlated with the types $\sigma_{i}$ of the items (see \cite{saade2016clustering}). We use a deflation-based power iteration method \cite{thang2013deflation} to approximate these eigenvectors, starting from informative initial conditions incorporating the knowledge drawn from the partially labeled data. Numerical simulations illustrating the performance of this algorithm are presented in section \ref{sec:numerical_simulations}. Note that each deflated matrix $\B_{c}$ for $c\geq 2$ is a rank-$(c-1)$ perturbation of a sparse matrix, so that the power iteration can be done efficiently using sparse linear algebra routines. In particular, both algorithms \ref{alg:2_clusters} and \ref{alg:q_clusters} have a time and space complexities linear in the number of items $n$.
\begin{algorithm}[]
\caption{Non-backtracking local walk ($q$ clusters)}
\label{alg:q_clusters}
\begin{algorithmic}[1]
\Require{$n\,,q\,,\mathcal{L}\,,(\sigma_i\in[q])_{i\in\mathcal{L}}\,,E,(w_{ij})_{(ij)\in E}\,,k_{\rm max}$}
\Ensure{Cluster assignments $(\hat{\sigma}_i)_{i\in[n]}$ }
\State $\B_{1}\leftarrow \B$ where $\B$ is the matrix of equation (\ref{def_B})
\For{$c=1,\cdots,q-1$} 
\State \textbf{Initialize} the messages $v^{(0)}=(v_{i\to j}^{(0)})_{(ij)\in E}$\
\Indent
\ForAll{$(i\to j)\in \vec{E}$} \unskip 
\LineIf{$i\in\mathcal{L}$ and $\sigma_i=c$}{$v_{i\to j}^{(0)} \leftarrow 1$}
\LineElsIf{$i\in\mathcal{L}$ and $\sigma_i\neq c$}{$v_{i\to j}^{(0)} \leftarrow -1$}
\LineElse{$v_{i\to j}^{(0)}\leftarrow \pm 1$ uniformly at random}
\EndFor
\EndIndent
\State \textbf{Iterate} for $k=1,\dots,k_{\rm max}$
\Indent
\State $v^{(k)}\leftarrow \B_{c} v^{(k-1)}$ 
\EndIndent
\State \textbf{Pool} the messages in a vector $\hat{v}_{c}\in\mathbb{R}^{n}$ with 
\State entries $(\hat{v}_{i,c})_{i\in[n]}$
\Indent
\LineFor{$i\in [n]$}{$\hat{v}_{i,c} \leftarrow \sum_{l\in\partial i} w_{il}v_{l\to i}^{(k_{\rm max})}$}
\EndIndent
\State \textbf{Deflate} $\B_{c}$
\Indent
\Indent
\State $\displaystyle \B_{c+1} \leftarrow \B_{c} - \frac{\B_{c}v^{(k_{\rm max})}\trans{v^{(k_{\rm max})}}\B_{c}}{\trans{v^{(k_{\rm max})}}\B_{c}v^{(k_{\rm max})}}$
\EndIndent
\EndIndent
\EndFor
\State \textbf{Concatenate} $\hat{V}\leftarrow [\hat{v}_{1}|\cdots |\hat{v}_{q-1}]\in\mathbb{R}^{n\times(q-1)}$ 
\State \textbf{Output} the assignments $(\hat{\sigma}_{i})_{i\in[n]}\leftarrow {\rm kmeans}(\hat{V})$
\end{algorithmic}
\end{algorithm}


\section{Numerical simulations}
\label{sec:numerical_simulations}

\begin{figure*}[h]
\begin{center}
\includegraphics[width=0.8\linewidth]{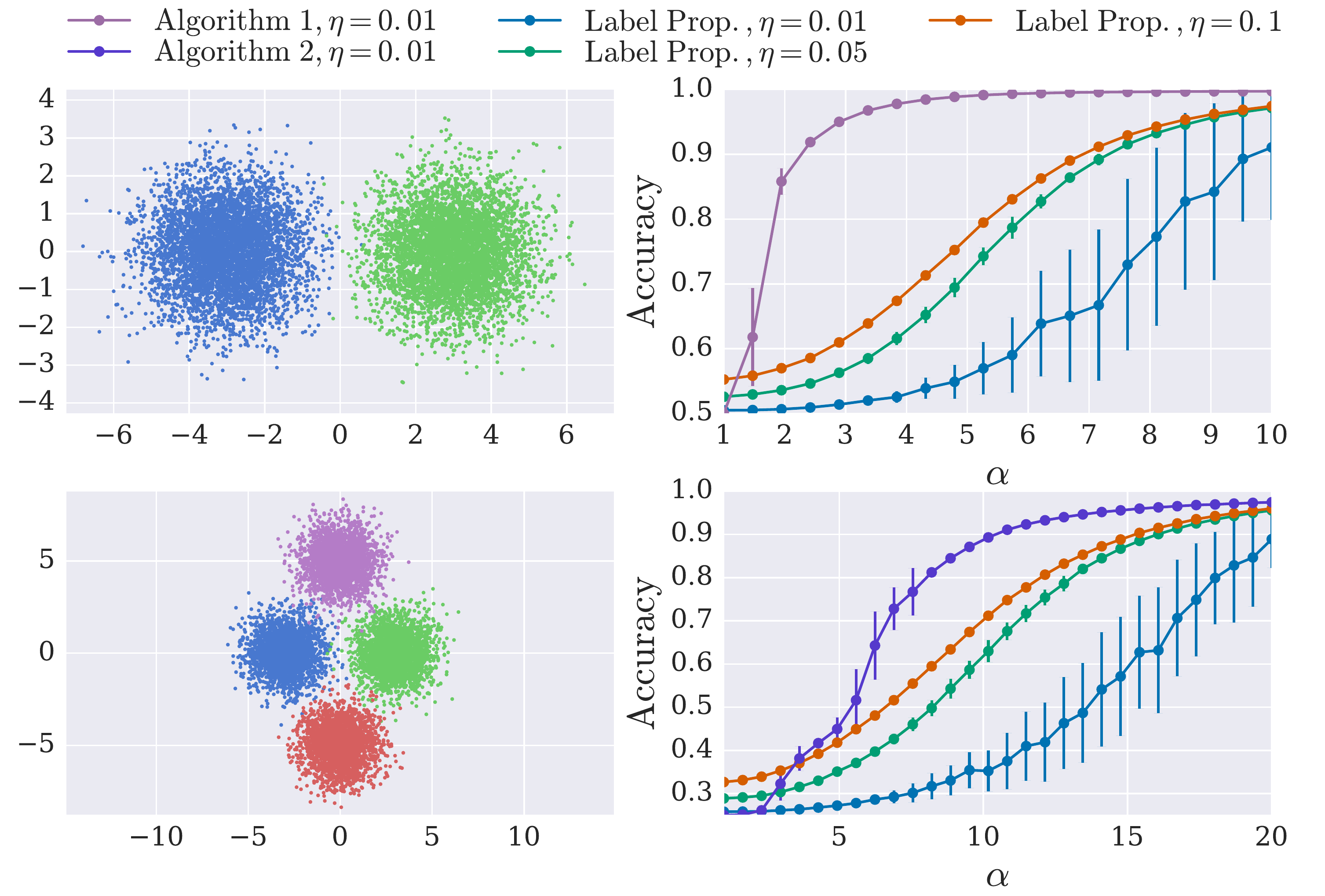}
\end{center}
\vspace{-1em}
\caption{ Performance of algorithms \ref{alg:2_clusters} and \ref{alg:q_clusters} compared to label propagation on a toy dataset in two dimensions. The left panel shows the data, composed of $n=10^{4}$ points, with their true labels. 
The right panel shows the clustering performance on random subsamples of the complete similarity graph.
Each point is averaged over $100$ realizations.
The accuracy is defined as the fraction of correctly labeled points. We set the maximum number of iterations of our algorithms to $k_{\rm max} = 30$. $\alpha$ is the average degree of the Erd\H{o}s-R\'enyi random graph $G$, and $\eta$ is the fraction of labeled data. For all methods, we used the same similarity function $s_{ij}=\exp{-d_{ij}^{2}/\sigma^{2}}$ where $d_{ij}$ is the Euclidean distance between points $i$ and $j$ and $\sigma^{2}$ is a scaling factor which we set to the empirical mean of the observed squared distances $d_{ij}^{2}$. For algorithms \ref{alg:2_clusters} and \ref{alg:q_clusters}, we used the weighting function $w(s):=s-\bar{s}$ (i.e. we simply center the similarities, see text). Label propagation is run on the random similarity graph $G$. We note that we improved the performance of label propagation by using only, for each point, the similarities between this point and its three nearest neighbors in the random graph $G$.
}    
\label{fig:blobs}
\end{figure*}

\begin{figure*}
\begin{center}
\includegraphics[width=1.0\linewidth]{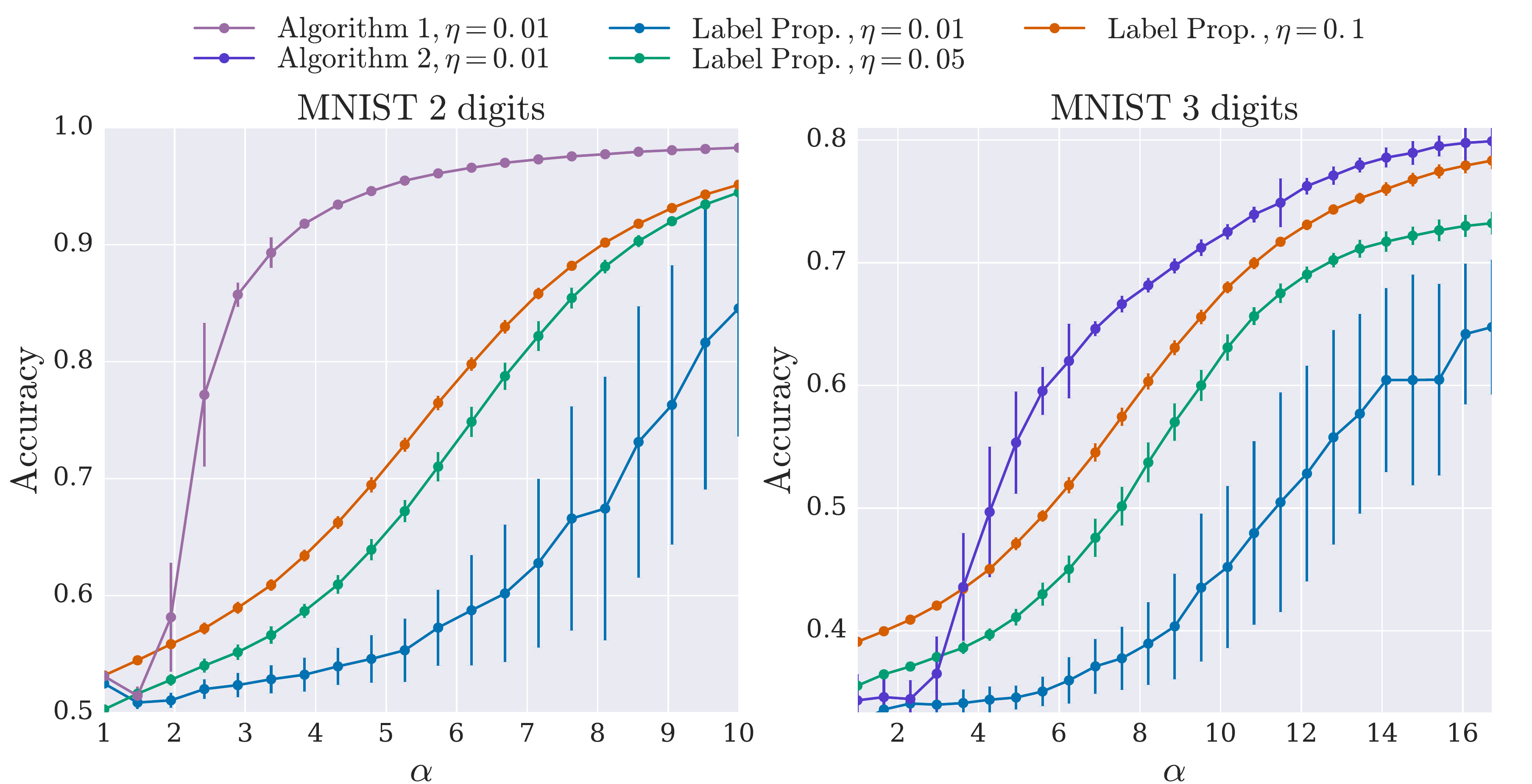}
\end{center}
\caption{ Performance of algorithms \ref{alg:2_clusters} and \ref{alg:q_clusters} compared to label propagation on 
a subset of the MNIST dataset. The left panel corresponds the set of 0's and 1's ($n=14780$ samples) while the right panel corresponds the 0's,1's and 2's ($n = 21770$). All the parameters are the same as in the caption of figure \ref{fig:blobs}, except that we used the Cosine distance in place of the Euclidean one. 
}    
\label{fig:MNIST}
\end{figure*}

\begin{figure*}[t]
\begin{center}
\includegraphics[width=0.65\linewidth]{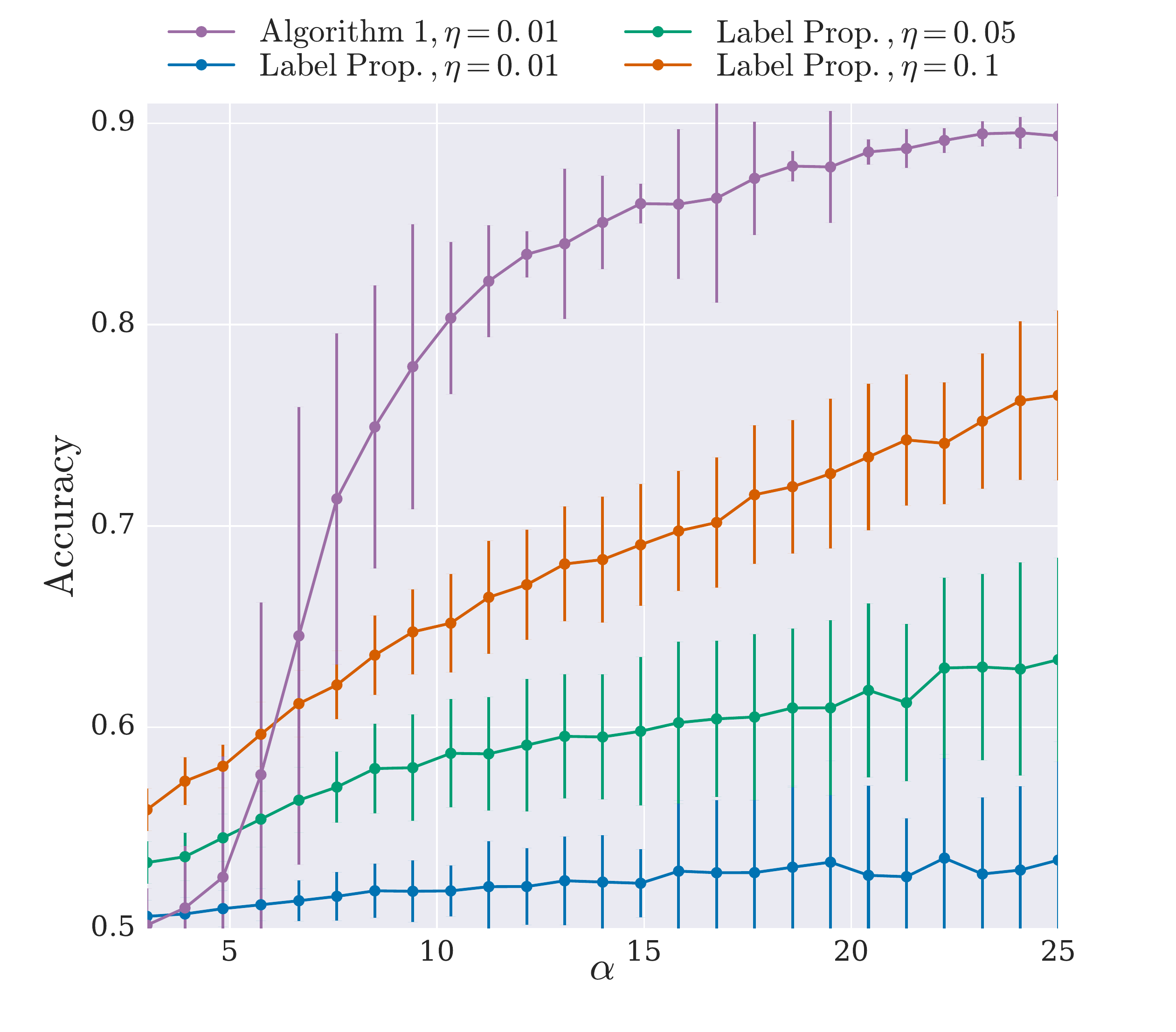}
\end{center}
\vspace{-2em}
\caption{ Performance of algorithm \ref{alg:2_clusters} compared to label propagation on a subset of the $20$ Newsgroups dataset with $q=2$  clusters. We consider the topics ``misc.forsale'' ($975$ documents) and `` soc.religion.christian'' ($997$ documents), which are relatively easy to distinguish, to illustrate the efficiency of our subsampling approach. The resulting dataset consists of $n = 1972$ text documents, to which we applied a standard ``tf-idf'' transformation (after stemming, and using word unigrams) to obtain a vector representation of the documents. We used the same similarity (based on the Cosine distance) and weighting functions as in figure~\ref{fig:MNIST}.  
}    
\label{fig:newsgroups}
\end{figure*}

In addition to the theoretical guarantees presented in the previous section, we ran numerical simulations on two toy datasets consisting of 2 and 4 Gaussian blobs (figure \ref{fig:blobs}), and two subsets of the MNIST dataset \cite{lecun1998gradient} consisting respectively of the digits in $\{0,1\}$ and $\{0,1,2\}$ (figure \ref{fig:MNIST}). We also considered the $20$ Newsgroups text dataset \cite{lang1995newsweeder}, consisting of text documents organized in $20$ topics, of which we selected $2$ for our experiments of figure~\ref{fig:newsgroups}.
All three examples differ considerably from the model we have studied analytically. In particular, the random similarities are not identically distributed conditioned on the true labels of the items, but depend on latent variables, such as the distance to the center of the Gaussian, in the case of figure \ref{fig:blobs}. Additionally, in the case of the MNIST dataset of figure \ref{fig:MNIST}, the clusters have different sizes (e.g. $6903$ for the 0's and $7877$ for the 1's).
Nevertheless, we find that our algorithm performs well, and outperforms the popular label propagation algorithm \cite{zhu2002learning} in a wide range of values of the sampling rate $\alpha$.

In all cases, we find that the accuracy achieved by algorithms \ref{alg:2_clusters} and \ref{alg:q_clusters} is an increasing function of $\alpha$, rapidly reaching a plateau at a limiting accuracy. Rather than the absolute value of this limiting accuracy, which depends on the choice of the similarity function, perhaps the most important observation is the rate of convergence of the accuracy to this limiting value, as a function of $\alpha$. Indeed, on these simple datasets, it is enough to compute, for each item, their similarity with a few randomly chosen other items to reach an accuracy within a few percents of the limiting accuracy allowed by the quality of the similarity function. As a consequence, similarity-based clustering can be significantly sped up. For example, we note that the semi-supervised clustering of the 0's and 1's of the MNIST dataset (representing $n=14780$ points in dimension $784$), from $1\%$ labeled data, and to an accuracy greater than $96\%$ requires $\alpha\approx 6$ (see figure \ref{fig:MNIST}), and runs on a laptop in 2 seconds, including the computation of the randomly chosen similarities. Additionally, in contrast with our algorithms, we find that in the strongly undersampled regime (small $\alpha$), the performance of label propagation depends strongly on the fraction $\eta$ of available labeled data. We find in particular that algorithms \ref{alg:2_clusters} and \ref{alg:q_clusters} outperform label propagation even starting from ten times fewer labeled data.

\section{Proof of theorem \ref{proposition}}
\label{sec:analysis_density_evolution}
Consider the model introduced in section \ref{sec:algo_guarante} for the case of two clusters. We will bound the probability of error on a randomly chosen node, and the different results will follow. 
Denote by $I$ an integer drawn uniformly at random from $[n]$, and by $\hat{\sigma}_{I}^{(k)} = \pm 1$ the decision variable after $k$ iterations of algorithm \ref{alg:2_clusters}.
We are interested in the probability of error at node $I$ conditioned on the true label of node $I$, i.e. $\P(\hat{\sigma}_{I}^{(k)} \neq \sigma_{I}|\sigma_I)$.
As noted previously, the algorithm is local in the sense that $\hat{\sigma}_{I}^{(k)}$ depends only
on the messages in the neighborhood of $I$ consisting of all the nodes and edges of $G$ that are at most $k+1$ steps aways from $I$. By bounding the total variation distance between the law of $G_{I,k+1}$ and a weighted Galton-Watson branching process, we show \cite[see][prop.~31]{blm15}
\begin{equation}
\label{P_error}
\P\left(\hat{\sigma}_{I}^{(k)} \neq \sigma_{I}|\sigma_I\right) \leq \P\left(\sigma_{I}\hat{v}_{\sigma_{I}}^{(k)} \leq 0\right) + C\frac{\alpha^{k+1}\log n}{\sqrt{n}}\, ,
\end{equation}
where $C>0$ is a constant, and the random variables $\hat{v}_{\sigma}^{(k)}$ for $\sigma = \pm 1$ are distributed according to
\begin{align}
\label{density_full}
\hat{v}_{\sigma}^{(k)} \overset{\mathcal{D}}{=} \sum_{i=1}^{d_1} w_{i,\rm in} v_{i,\sigma}^{(k)} + \sum_{i=1}^{d_2} w_{i,\rm out} v_{i,-\sigma}^{(k)}\, ,
\end{align}
where $\overset{\mathcal{D}}{=}$ denotes equality in distribution. The random variables $v_{i,\sigma}^{(k)}$ for $\sigma=\pm 1$ have the same distribution as the message $v_{i\to j}^{(k)}$ after $k$ iterations of the algorithm, for a randomly chosen edge $(i\to j)$, conditioned on the type of node $i$ being $\sigma$. They are i.i.d. copies of a random variable $v_{\sigma}^{(k)}$ whose distribution is defined recursively, for $l \geq 0$ and $\sigma=\pm 1$, through 
\begin{align}
\label{density_cav}
v_{\sigma}^{(l+1)} \overset{\mathcal{D}}{=} \sum_{i=1}^{d_1} w_{i,\rm in} v_{i,\sigma}^{(l)} + \sum_{i=1}^{d_2} w_{i,\rm out} v_{i,-\sigma}^{(l)}\, .
\end{align} 
In equations (\ref{density_full}), $d_1$ and $d_2$ are two independent random variables with a Poisson distribution of mean $\alpha/2$, and $w_{i,\rm in}$ (resp. $w_{i,\rm out}$) are i.i.d copies of $w_{\rm in}$ (resp. $w_{\rm out}$) whose distribution is the same as the weights $w_{ij}$, conditioned on $\sigma_i = \sigma_j$ (resp. $\sigma_i \neq \sigma_j$). Note in particular that $\hat{v}_{\sigma}^{(k)}$  has the same distribution as~$v_{\sigma}^{(k+1)}$. 

Theorem \ref{proposition} will follow by analyzing the evolution of the first and second moments of the distribution of $v_{+1}^{(k+1)}$ and $v_{-1}^{(k+1)}$.
Equations (\ref{density_cav}) can be used to derive recursive formulas for the first and second moments. In particular, the expected values verify the following linear system
\begin{equation}
\label{matrix_recursion}
\begin{pmatrix}[2.0]
\E\left[v_{+1}^{(l+1)}\right]  \\
\E\left[v_{-1}^{(l+1)}\right] 
\end{pmatrix} 
= \frac{\alpha}{2}
\begin{pmatrix}[2.0]
\E[w_{\rm in}] & \E[w_{\rm out}] \\
\E[w_{\rm out}] & \E[w_{\rm in}]
\end{pmatrix} 
\begin{pmatrix}[2.0]
\E\left[v_{+1}^{(l)}\right]  \\
\E\left[v_{-1}^{(l)}\right] 
\end{pmatrix}\, .
\end{equation}
The eigenvalues of this matrix are $\E\left[w_{\rm in}\right] + \E\left[w_{\rm out}\right]$ with eigenvector $\trans{(1,1)}$, and $\E\left[w_{\rm in}\right] - \E\left[w_{\rm out}\right]$ with eigenvector $\trans{(1,-1)}$. With the assumption of our model, we have 
$\E\left[v_{+1}^{(0)}\right] = \eta = - \E\left[v_{-1}^{(0)}\right]$ 
which is proportional to the second eigenvector. 
Recalling the definition of $\Delta(w)$ from section \ref{sec:algo_guarante}, we therefore have, for any $l\geq 0$,
\begin{equation}
\label{recurrence_first_moment}
\E\left[v_{+1}^{(l+1)}\right] = \alpha \Delta(w) \E\left[v_{+1}^{(l)}\right]\, ,
\end{equation}
and $\E\left[v_{-1}^{(l)}\right] = - \E\left[v_{+1}^{(l)}\right]$. With the additional observation that $\E\left[\left(v_{+1}^{(0)}\right)^{2}\right] = \E\left[\left(v_{-1}^{(0)}\right)^{2}\right] = 1$, a simple induction shows that for any $l\geq 0$, 
\begin{equation}
\E\left[\left(v_{+1}^{(l)}\right)^{2}\right] = \E\left[\left(v_{-1}^{(l)}\right)^{2}\right]\, .
\end{equation}
Recalling the definition of $\Sigma(w)^{2}$ from section \ref{sec:algo_guarante}, we have the recursion 
\begin{equation}
\begin{aligned}
\label{recurrence_second_moment}
\E\left[\left(v_{+1}^{(l+1)}\right)^{2}\right] =\, \alpha^{2}\Delta(w)^{2}\, \E\left[v_{+1}^{(l)}\right]^2 +\, \alpha\,\Sigma(w)^{2}\,\E\left[\left(v_{+1}^{(l)}\right)^{2}\right]\, .
\end{aligned}
\end{equation}
Noting that since $\Delta(w) > 0$, we have $\sigma\, \E\left[v_{\sigma}^{(k+1)}\right] > 0$ for $\sigma = \pm 1$, the proof of theorem \ref{proposition} is concluded by  invoking Cantelli's inequality
\begin{equation}
\P\left(\sigma\,{v}_{\sigma}^{(k+1)} \leq 0\right)  \leq 1 - r_{k+1}\, , 
\end{equation}
with, for $l \geq 0$,
\begin{equation}
r_{l} := \E\left[v_{\sigma}^{(l)}\right]^{2}\E\left[\left(v_{\sigma}^{(l)}\right)^{2}\right]^{-1} \, ,
\end{equation}
where $r_{l}$ is independent of $\sigma$, and is shown to verify the recursion (\ref{proposition_rec_r}) by combining (\ref{recurrence_first_moment}) and (\ref{recurrence_second_moment}).


\section{Proof of theorem \ref{theorem}}
\label{proof_theorem}
The proof is adapted from a technique developed by \cite{karger2011iterative}. We show that the random variables $v_{\sigma}^{(l)}$ are sub-exponential by induction on $l$. A random variable $X$ is said to be sub-exponential if there exist constants $K>0,a,b$ such that for $|\lambda|<K$
\begin{equation}
\E[e^{\lambda X}] \leq \exp\left(\lambda a+\lambda^{2}b\right)\, .
\end{equation}
Define $f_{\sigma}^{(l)}(\lambda):=\E\left[e^{\lambda v_{\sigma}^{(l)}}\right]$ for $l\geq 0$ and $\sigma = \pm 1$. We introduce two sequences $(a_{l})_{l\geq 0},(b_{l})_{l\geq 0}$ defined recursively by $a_{0}=\eta,b_{0}=1/2$ and for $l\geq 0$
\begin{equation}
\begin{aligned}
\label{rec_a_b}
a_{l+1} &= \alpha\Delta(w) a_{l}\, ,\\
b_{l+1} &= \alpha\Sigma(w)^{2}\left(b_{l} + \frac{3}{2}\max(a_{l}^{2},b_{l})\right)\, .
\end{aligned}
\end{equation}
Note that since we assume that $\alpha\,\Delta(w)>1$ and $\alpha\,\Sigma(w)^{2}>1$, both sequences are positive and increasing. In the following, we show that 
\begin{align}
\label{recurrence_f}
f_{\sigma}^{(k+1)}(\lambda) \leq \exp\left(\sigma\lambda a_{k+1} + \lambda^{2} b_{k+1}\right)\, ,
\end{align}
for $|\lambda|\leq \left(2\max\left(a_{k},\sqrt{b_{k}}\right)\right)^{-1}$. Theorem \ref{theorem} will follow from the Chernoff bound applied at 
\begin{equation}
\lambda_{\sigma}^{*} = -\sigma \frac{a_{k+1}}{2b_{k+1}}\min\left(1,\frac{\Sigma(w)^{2}}{\Delta(w)}\right)\, . 
\end{equation}
The fact that $|\lambda_{\sigma}^{*}|\leq \left(2\max\left(a_{k},\sqrt{b_{k}}\right)\right)^{-1}$ follows from~(\ref{rec_a_b}). Noting that $\sigma \lambda_{\sigma}^{*} < 0$ for $\sigma=\pm 1$, the Chernoff bound allows to show 
\begin{equation}
\begin{aligned}
\P\left(\sigma v_{\sigma}^{(k+1)} \leq 0\right) &\leq f_{\sigma}^{(k+1)}\left(\lambda_{\sigma}^{*}\right)\\
&\leq \exp{\left[-\frac{q_{k+1}}{4}\min\left(1,\frac{\Sigma(w)^{2}}{\Delta(w)}\right)\right]}\, ,
\end{aligned}
\end{equation}
where $q_{k+1} := a_{k+1}^{2}/b_{k+1}$ is shown using (\ref{rec_a_b}) to verify the recursion (\ref{recurrence_q}). We are left to show that $f_{\sigma}^{(k+1)}(\lambda)$ verifies (\ref{recurrence_f}). First, with the choice of initialization in algorithm \ref{alg:2_clusters}, we have for any $\lambda\in\mathbb{R}$
\begin{equation}
\begin{aligned}
f_{+1}^{(0)}(\lambda) = f_{-1}^{(0)}(-\lambda) &= \frac{1+\eta}{2}\exp\left(\lambda\right) + \frac{1-\eta}{2}\exp\left(-\lambda\right)\\
 &\leq \exp\left(\eta\lambda + \lambda^{2}/2\right)\, , 
\end{aligned}
\end{equation}
where we have used the inequality for $x\in[0,1],\lambda\in\mathbb{R}$
\begin{equation}
xe^{\lambda}+(1-x)e^{-\lambda}\leq \exp\left((2x-1)\lambda+\lambda^{2}/2\right)\, .
\end{equation}
Therefore we have shown $f_{\sigma}^{(0)}(\lambda)\leq \exp{(\sigma\lambda a_{0} + \lambda^{2}b_{0})}$. Next, let us assume that for some $l\geq 0$ and for any $\lambda$ such that $|\lambda|\leq \left(2\max\left(a_{l-1},\sqrt{b_{l-1}}\right)\right)^{-1}$, 
\begin{equation}
f_{\sigma}^{(l)}(\lambda)\leq \exp{(\sigma\lambda a_{l} + \lambda^{2}b_{l})}\, ,
\end{equation}
with the convention $a_{-1}=b_{-1}=0$ so that the previous statement is true for any $\lambda\in\mathbb{R}$ if $l=0$. The density evolution equations (\ref{density_cav}) imply the following recursion on the moment-generating functions, for any $\lambda\in\mathbb{R},\sigma=\pm1$, 
\begin{equation}
\begin{aligned}
\label{recurrence_generating_function}
f_{\sigma}^{(l+1)}(\lambda) = \exp\Big( -\alpha + \frac{\alpha}{2}\Big( \E_{w_{\rm in}}\left[ f_{\sigma}^{(l)}(\lambda w_{\rm in})\right] 
+ \E_{w_{\rm out}}\left[ f_{-\sigma}^{(l)}(\lambda w_{\rm out})\right] \Big)  \Big)\, .
\end{aligned}
\end{equation}
We claim that for $|\lambda|\leq \left(2\max\left(a_{l},\sqrt{b_{l}}\right)\right)^{-1}$ and for $\sigma=\pm1$,
\begin{equation}
\begin{aligned}
\label{final_claim}
\frac{1}{2}\left( \E_{w_{\rm in}}\left[ f_{\sigma}^{(l)}(\lambda w_{\rm in})\right] + \E_{w_{\rm out}}\left[ f_{-\sigma}^{(l)}(\lambda w_{\rm out})\right] \right) 
\leq 1 + \sigma a_{l}\Delta(w)\lambda + \lambda^{2}\Sigma(w)^{2}\left( b_{l} + \frac{3}{2}\max(a_{l}^{2},b_{l}) \right)
\end{aligned}
\end{equation}
Injecting equation (\ref{final_claim}) in the recursion (\ref{recurrence_generating_function}) yields $f_{\sigma}^{(l+1)}(\lambda)\leq \exp{(\sigma\lambda\, a_{l+1} + \lambda^{2}\,b_{l+1})}$, for any $\lambda$ such that $|\lambda|\leq \left(2\max\left(a_{l},\sqrt{b_{l}}\right)\right)^{-1}$, with $a_{l+1},b_{l+1}$ defined by (\ref{rec_a_b}). The proof is then concluded by induction on $0\leq l \leq k$.
To show (\ref{final_claim}), we start from the following inequality: for $|a|\leq 3/4$,
\begin{equation}
\exp(a)\leq 1 + a + (2/3)a^{2}\, .
\end{equation}
With $|w|\leq 1$ as per the assumption of theorem \ref{theorem}, for $|\lambda|\leq \left(2\max\left(a_{l},\sqrt{b_{l}}\right)\right)^{-1}$, we have for $\sigma=\pm1$ that $|\sigma\lambda w a_{l}  + \lambda^{2} w^{2} b_{l}| \leq 3/4$. Additionally, since $a_{l}$ and $b_{l}$ are non-decreasing in $l$, we also have that $|\lambda|\leq \left(2\max\left(a_{l-1},\sqrt{b_{l-1}}\right)\right)^{-1}$, so that by our induction hypothesis, for $\sigma = \pm1$
\begin{align}
&f_{\sigma}^{(l)}(\lambda w) \leq \exp\left(\sigma\lambda w a_{l}  + \lambda^{2} w^{2} b_{l}\right) \\
&\leq 1 + \sigma\lambda w a_{l}  + \lambda^{2} w^{2} b_{l} + \frac{2}{3}\left(\sigma\lambda w a_{l}  + \lambda^{2} w^{2} b_{l}\right)^{2} \\
&\leq 1 + \sigma\lambda w a_{l}  + \lambda^{2} w^{2} b_{l} + \frac{2}{3}\lambda^{2} w^{2}\left( a_{l}  + |\lambda| b_{l}\right)^{2} \\
&\leq 1 + \sigma\lambda w a_{l}  + \lambda^{2} w^{2} b_{l} + \frac{3}{2}\lambda^{2} w^{2}\max(a_{l}^{2},b_{l})\, ,
\end{align}
where we have used in the last inequality that $\left( a_{l}  + |\lambda| b_{l}\right)^{2}\leq 9/4\max(a_{l}^{2},b_{l})$. The claim (\ref{final_claim}) follows by taking expectations, and the proof is completed.

\section*{Acknowledgement}
  This work has been supported by the ERC under the European Union’s FP7 Grant Agreement 307087-SPARCS and by the French Agence Nationale de la Recherche under reference ANR-11-JS02-005-01 (GAP project).

\bibliographystyle{IEEEtran}
\bibliography{mybib}



\end{document}